\documentclass[conference,a4paper]{IEEEtran}
\IEEEoverridecommandlockouts

\usepackage[hidelinks]{hyperref}
\usepackage[cmex10]{amsmath}
\usepackage{amssymb,amsfonts}
\usepackage{dblfloatfix}
\usepackage{multirow} 

\usepackage[ruled,vlined]{algorithm2e}
\usepackage{graphicx}
\graphicspath{{Figures/PDF/}{Figures/PNG/}}

\usepackage{booktabs}
\usepackage{siunitx}
\usepackage[numbers,compress]{natbib}
\usepackage{texnames}
\usepackage{bm,bbm}
\usepackage{orcidlink}

\begin{document}

\title{\fontsize{23}{25}\selectfont\uppercase{Uncertainty-Aware Test-Time Adaptation for Cross-Region Spatio-Temporal Fusion of Land Surface Temperature}
\thanks{This work was supported by Orléans Métropole and Région Centre-Val de Loire (Corresponding author*: Sofiane Bouaziz).

Copyright 2026 IEEE. Published in the 2026 IEEE International Geoscience and Remote Sensing Symposium (IGARSS 2026), scheduled for 9 - 14 August 2026 in Washington, D.C.. Personal use of this material is permitted. However, permission to reprint/republish this material for advertising or promotional purposes or for creating new collective works for resale or redistribution to servers or lists, or to reuse any copyrighted component of this work in other works, must be obtained from the IEEE. Contact: Manager, Copyrights and Permissions / IEEE Service Center / 445 Hoes Lane / P.O. Box 1331 / Piscataway, NJ 08855-1331, USA. Telephone: + Intl. 908-562-3966.}
}

\author{ \IEEEauthorblockN{
Sofiane Bouaziz\textsuperscript{1,2,*}, 
Adel Hafiane\textsuperscript{2}, 
Raphaël Canals\textsuperscript{2}, 
Rachid Nedjai\textsuperscript{3}
}

\IEEEauthorblockA{\textsuperscript{1}INSA CVL, Université d’Orléans, PRISME UR 4229, Bourges, 18022, Centre Val de Loire, France}

\IEEEauthorblockA{\textsuperscript{2}Université d’Orléans, INSA CVL, PRISME UR 4229, Orléans, 45067, Centre Val de Loire, France}

\IEEEauthorblockA{\textsuperscript{3}Université d'Orléans, CEDETE, UR 1210, Orléans, 45067, Centre Val de Loire, France}
}

\maketitle
\begin{abstract}
Deep learning models have shown great promise in diverse remote sensing applications. However, they often struggle to generalize across geographic regions unseen during training due to domain shifts. Domain shifts occur when data distributions differ between the training region and new target regions, due to variations in land cover, climate, and environmental conditions. Test-time adaptation (TTA) has emerged as a solution to such shifts, but existing methods are primarily designed for classification and are not directly applicable to regression tasks. In this work, we address the regression task of spatio-temporal fusion (STF) for land surface temperature estimation. We propose an uncertainty-aware TTA framework that updates only the fusion module of a pre-trained STF model, guided by epistemic uncertainty, land use and land cover consistency, and bias correction, without requiring source data or labeled target samples. Experiments on four target regions with diverse climates, namely Rome in Italy, Cairo in Egypt, Madrid in Spain, and Montpellier in France, show consistent improvements in RMSE and MAE for a pre-trained model in Orléans, France. The average gains are 24.2\% and 27.9\%, respectively, even with limited unlabeled target data and only 10 TTA epochs.
\end{abstract}


\section{Introduction}
\begin{figure}[h]
  \centering
  \includegraphics[width=0.45\textwidth]
  {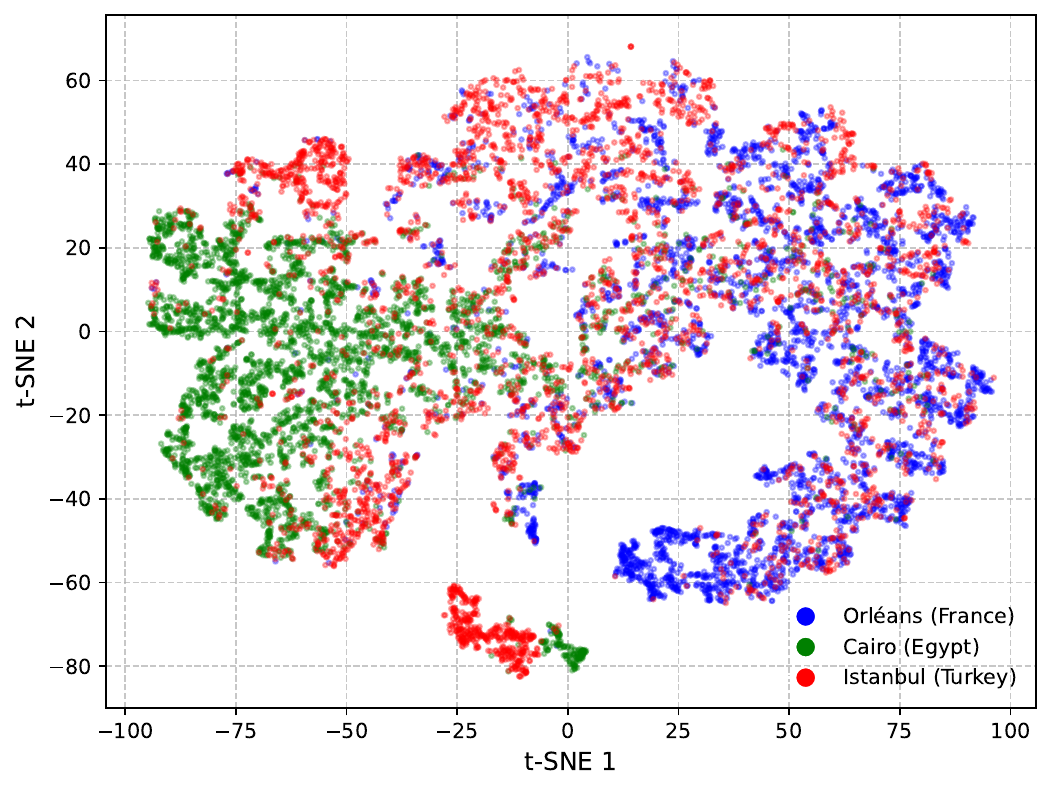}
\caption{t-SNE visualization of LULC distributions for Orléans in France, Cairo in Egypt, and Istanbul in Turkey, derived from Landsat 8 observations acquired on 04th, 05th, and 08th Aug 2025, respectively. The embedding is constructed using the NDVI, NDWI, and NDBI spectral indices.}

  \label{fig:tsne_plot}
\end{figure}

Deep learning (DL) has recently driven major progress in remote sensing (RS), with successful applications including semantic segmentation~\cite{huang2023deep, deng2021semantic}, change detection~\cite{gui2024remote, napiorkowska2018three}, disaster monitoring~\cite{al2024integrating, 11242903}, and spatio-temporal fusion (STF)~\cite{belgiu2019spatiotemporal, song2020remote}. However, most of these approaches operate under the assumption that both training (source domain) and test data (target domain) are drawn independently and identically from the same distribution~\cite{quinonero2008dataset}. In real-world Earth observation scenarios, this assumption rarely holds, as DL models are often applied across different geographic regions, acquisition settings, and environmental conditions than those seen during their training~\cite{tuia2016domain}. Such differences lead to the domain shift problem, which result in significant performance degradation~\cite{liang2025comprehensive}. This issue is especially pronounced for land surface temperature (LST), whose spatial patterns strongly depend on climate, land cover, and urban structure, making generalization across regions particularly challenging~\cite{bouaziz2024deep}. Fig.~\ref{fig:tsne_plot} presents a t-distributed stochastic neighbor embedding (t-SNE) visualization of land use and land cover (LULC) based on three spectral indices derived from Landsat 8, namely the normalized difference vegetation index (NDVI), the normalized difference water index (NDWI), and the normalized difference built-up index (NDBI), for three geographically distinct regions, Orléans in France, Cairo in Egypt, and Istanbul in Turkey. The embedding reveals a clear clustering of samples for each region. This illustrates the domain shift problem, where a model trained on Orléans is likely to struggle when applied to regions with different LULC compositions, such as Cairo.

\vspace{0.2em}
To mitigate the impact of domain shift, transfer learning (TL) has emerged as a widely adopted strategy, aiming to reuse knowledge learned from the training source domain to improve performance on an unseen target domain~\cite{ma2024transfer}. Existing TL approaches take several forms. Unsupervised domain adaptation exploits labeled data from the source domain to align feature distributions and learn a model that generalizes to an unlabeled target domain~\cite{liu2022deep}. However, this setting assumes continued access to source domain data, which is often impractical in RS due to data confidentiality, storage constraints, or limited data availability~\cite{wang2024test, cao2023multi}. Fine-tuning approaches instead adapt a pre-trained model by updating part or all of its parameters using labeled samples from the target domain~\cite{yosinski2014transferable}, which are often difficult to acquire.  More recently, test-time adaptation (TTA) has been proposed as an alternative paradigm that eliminates the need for both source domain data and labeled target samples, and instead adapts a pre-trained model directly using only unlabeled target data prior to inference~\cite{liang2025comprehensive}.

\vspace{0.2em}

Despite these advances, most TTA methods have been developed for classification tasks~\cite{liang2025comprehensive}, where unsupervised proxy objectives such as entropy minimization~\cite{wang2020tent} and mutual information maximization~\cite{liang2021source} can guide the pre-trained model adaptation. These metrics are effective as classification models produce predictive probability distributions that can be directly optimized. Extending TTA to regression is less straightforward, even though regression is one of the most common tasks in DL and RS~\cite{adachi2024test}. Standard classification-based metrics cannot be applied, since regression models output only scalar values rather than predictive distributions~\cite{adachi2024test}.

\vspace{0.2em}

In this paper, we focus on STF for LST estimation as a regression pre-trained model and propose an uncertainty-aware TTA method to extend its applicability across different regions worldwide. To the best of our knowledge, this is the first TTA method specifically designed for regression tasks in RS. Our key contributions are as follows:

\begin{itemize}
    \item We propose an unsupervised loss that integrates epistemic uncertainty and LULC correlations to guide TTA.
    \item We introduce a partial weight update strategy for STF by freezing most of the model parameters and updating only those responsible for feature space fusion.
    \item We demonstrate the effectiveness of our approach on four different regions with minimal TTA training epochs.
\end{itemize}

\section{Related works}
Our work builds upon recent advances in TTA by introducing an uncertainty-aware framework for STF of LST.

\vspace{0.2em}
\noindent \textit{Test-Time Adaptation} adapts a pre-trained model to the target domain using only unlabeled target data, without requiring access to the source domain~\cite{liang2025comprehensive}. Unlike traditional domain adaptation methods, TTA operates strictly at inference time~\cite{xiao2024beyond}. Existing TTA approaches have been primarily developed for classification tasks, where objectives such as entropy minimization and mutual information maximization can effectively guide adaptation~\cite{liang2025comprehensive}, including in RS applications~\cite{11314052, huang2024learning}. However, extending these strategies to regression problems remains challenging, as regression models do not output predictive distributions~\cite{adachi2024test}.

\vspace{0.2em}
\noindent \textit{Uncertainty Estimation} is a crucial component of DL models, particularly in safety-critical applications~\cite{upadhyay2021uncertainty, sudarshan2021towards}, as it identifies unreliable predictions that can trigger corrective actions when model confidence is low~\cite{gawlikowski2023survey}. Predictive uncertainty is commonly decomposed into aleatoric uncertainty, which reflects noise in the data, and epistemic uncertainty, which captures uncertainty in the model parameters~\cite{he2025survey}. Bayesian methods provide a mathematical framework for modeling epistemic uncertainty~\cite{blundell2015weight, mobiny2021dropconnect}, but their direct application to DL networks is computationally prohibitive~\cite{gal2016dropout}. As a result, practical approximations such as Monte Carlo (MC) dropout are widely used to estimate epistemic uncertainty via stochastic forward passes~\cite{gal2016dropout, srivastava2014dropout}.

\vspace{0.2em}
\noindent\textit{Spatio-temporal Fusion of Land Surface Temperature} aims to generate LST estimates with both high spatial and temporal resolution by integrating observations from multiple satellite sensors~\cite{bouaziz2024deep}. Such information is critical for public health monitoring~\cite{estrela2023remote} and climate adaptation~\cite{sirmacek2022remote}. WGAST~\cite{bouaziz2025wgast} is a recent STF method that produces daily LST estimates at a spatial resolution of 10~m by relying on Terra MODIS 1~km LST at the target time, together with Terra MODIS 1~km LST, Landsat~8 30~m LST and LULC, and Sentinel-2 10~m LULC information acquired at a previous reference time. In this work, WGAST is adopted as the pre-trained STF model and serves as the baseline for our cross-region TTA framework.

\section{Methodology}

\subsection{Overview}

\begin{figure}[h]
  \centering
  \includegraphics[width=0.48\textwidth]
  {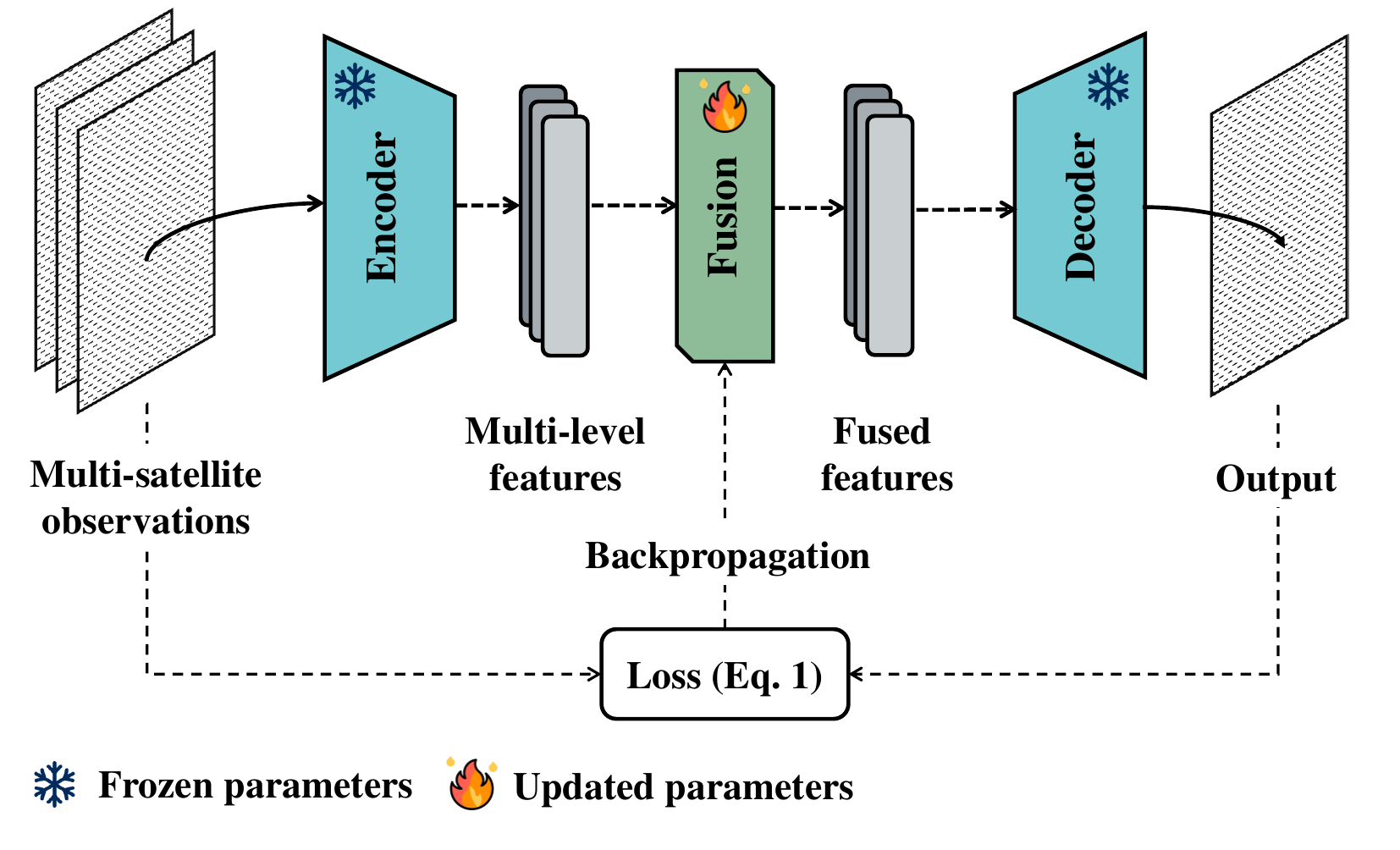}
\caption{Overview of the proposed TTA framework applied to the EFD architecture used in WGAST~\cite{bouaziz2025wgast}. The encoder and decoder parameters are frozen during TTA, and only the fusion module is updated according to the loss defined in Eq.~\ref{eq:ltta}.}
\label{figure:EDA}
\end{figure}

STF models generally adopt an encoder-fusion-decoder (EFD) architecture, as illustrated in Fig.~\ref{figure:EDA}. These models typically differ in the design of the encoder and decoder, the fusion mechanism, and the training strategy. WGAST~\cite{bouaziz2025wgast} employs a generator with an EFD structure, where the encoder and decoder consist of convolutional and deconvolutional layers with downsampling and residual blocks, and the model is trained using adversarial learning. In this work, we freeze the encoder and decoder parameters and update only the fusion module. The parameters of the fusion module are updated according to the loss function defined in Eq.~\ref{eq:ltta}.
\begin{equation}
\mathcal{L}_{\text{TTA}}(\hat{Y}) = \lambda_1 \mathcal{L}_{\text{uncertainty}}(\hat{Y}) 
+ \lambda_2 \mathcal{L}_{\text{LULC}}(\hat{Y}, I) 
+ \lambda_3 \mathcal{L}_{\text{bias}}(\hat{Y}, X)
\label{eq:ltta}
\end{equation}

\noindent where $\hat{Y}$ denotes the 10~m LST predicted by WGAST at the target time $t_2$, $I$ represents the LULC characteristics (NDVI, NDWI, and NDBI) at a prior reference time $t_1$, and $X$ is the Terra MODIS 1~km LST at $t_2$. The coefficients $\lambda_1$, $\lambda_2$, and $\lambda_3$ are weighting parameters that balance the contributions of each term. The overall TTA objective is obtained by aggregating the loss over all unlabeled target samples, as defined in Eq.~\ref{eq:tta_total}.
\begin{equation}
\mathcal{L}_{\text{TTA}} = \frac{1}{T} \sum_{i=1}^{T} \mathcal{L}_{\text{LTTA}}(\hat{Y}^{(i)})
\label{eq:tta_total}
\end{equation}

\noindent where $T$ denotes the number of samples in the target domain. Each component of this loss function is described in the following subsections. 

\subsection{Uncertainty-Aware Loss}
We propose an uncertainty-aware loss that discourages high epistemic uncertainty in the predicted LST. Epistemic uncertainty reflects the model’s lack of confidence arising from limited knowledge of the target domain and is particularly pronounced when the pre-trained STF model encounters unseen spatial or climatic conditions. Minimizing this uncertainty during TTA encourages the model to adjust its parameters toward more confident and stable predictions on the target domain. This makes epistemic uncertainty a suitable metric for guiding regression-based TTA.

\vspace{0.2em}
Epistemic uncertainty is estimated using MC dropout by enabling dropout layers at inference time and performing $N$ stochastic forward passes through the pre-trained STF model. Given the set of predictions $\{\hat{Y}_i\}_{i=1}^{N}$, the pixel-wise epistemic uncertainty is computed as the variance across MC samples, defined in Eq.~\ref{eq:epistemic}.
\begin{equation}
\text{Var}[\hat{Y}] = \frac{1}{N-1} \sum_{i=1}^{N} (\hat{Y}_i - \bar{\hat{Y}})^2, \quad
\bar{\hat{Y}} = \frac{1}{N} \sum_{i=1}^{N} \hat{Y}_i
\label{eq:epistemic}
\end{equation}

\vspace{0.2em}
The uncertainty-aware loss is then obtained by averaging the pixel-wise variance over the spatial dimensions of the predicted LST image, as shown in Eq.~\ref{eq:lcertainity}.
\begin{equation}
\mathcal{L}_{\text{uncertainty}}(\hat{Y})  = \frac{1}{HW} \sum_{i=1}^{H} \sum_{j=1}^{W} \text{Var}[\hat{Y}_{b,i,j}]
\label{eq:lcertainity}
\end{equation}

\noindent where $H$ and $W$ denote the height and width of the predicted LST image, respectively.

\subsection{Land use and land cover consistency Loss}

We introduce a LULC consistency loss that enforces physically meaningful relationships between the predicted LST and LULC characteristics, namely NDVI, NDWI, and NDBI. These indices are not direct physical measurements of LST but capture surface characteristics known to influence its dynamics. Moreover, since LULC patterns typically evolve slowly over time compared to short-term LST variations, the indices observed at the reference time $t_1$ provide reliable constraints for LST predictions at the target time $t_2$. Therefore, we compute the Pearson correlation coefficient between the predicted 10~m LST image ($\hat{Y}$) and each LULC index ($I_k$), after mean removal, as defined in Eq.~\ref{eq:pearson}.
\begin{equation}
\rho_k =
\frac{\sum (\hat{Y} - \mu_{\hat{Y}})(I_k - \mu_{I_k})}
{\sqrt{\sum (\hat{Y} - \mu_{\hat{Y}})^2 \sum (I_k - \mu_{I_k})^2}},
\label{eq:pearson}
\end{equation}

\vspace{0.2em}
Rather than enforcing index-specific correlations, we encourage the overall LST-LULC relationship. The LULC consistency loss is therefore defined as the average penalty over the absolute correlations across all indices, as shown in Eq.~\ref{eq:lphysics}.
\begin{equation}
\mathcal{L}_{\text{LULC}}(\hat{Y}, I) =
\frac{1}{3} \sum_{k=1}^{3} \left( 1 - |\rho_k| \right).
\label{eq:lphysics}
\end{equation}

\subsection{Bias Consistency Loss}

We propose a bias consistency loss based on first-order statistics to correct large-scale radiometric discrepancies between the predicted 10~m LST ($\hat{Y}$) and the Terra MODIS 1~km LST ($X$) at target time $t_2$, as defined in Eq.~\ref{eq:bias}.
\begin{equation}
\mathcal{L}_{\text{bias}}(\hat{Y}, X) = \left| \mu(\hat{Y}) - \mu(X) \right|
\label{eq:bias}
\end{equation}

\noindent where $\mu(\cdot)$ denotes the spatial average. This loss enables fast TTA while preserving the local spatial structures learned by the pre-trained STF model.

\subsection{Fusion Module Weight Update}

The encoder and decoder of STF models aim to project multi-satellite inputs into a latent representation and reconstruct the output. Their weights capture general feature representations and remain consistent across regions. In contrast, the fusion module is responsible for integrating the multi-source features and is more sensitive to regional variations. Therefore, during TTA, we optimize only the fusion parameters ($\mathbf{w}_{\text{fusion}}$) using the Adam optimizer with a learning rate $\eta$. The update at iteration $t$ is defined as in Eq.~\ref{eq:update}.
\begin{equation}
\mathbf{w}_{\text{fusion}}^{t+1} = \mathbf{w}_{\text{fusion}}^{t} - \eta \, \nabla_{\mathbf{w}_{\text{fusion}}} \mathcal{L}_{\text{TTA}}
\label{eq:update}
\end{equation}

\section{Experimental results}
\subsection{Experimental Settings}

\subsubsection{Data}

WGAST is pre-trained on data from Orléans (France)~\cite{bouaziz2025wgast}. We therefore evaluate its transferability using the proposed TTA framework on four geographically and climatically distinct target regions: Rome (Italy), Cairo (Egypt), Madrid (Spain), and Montpellier (France). These regions cover a wide range of climatic conditions, from Mediterranean (Rome, Montpellier) and continental Mediterranean (Madrid) to arid desert environments (Cairo). For each region, a limited number of target dates $t_2$ is selected, for which no high-resolution 10 m LST observations are available at inference time. The selected dates are summarized in Table~\ref{tab:target_dates}. Varying the number of target samples across regions allows us to evaluate both the adaptability and robustness of the proposed TTA method under limited unlabeled target data.

\setlength{\tabcolsep}{6pt}
\begin{table}[h]
\centering
\caption{Selected target dates $t_2$ for testing the transferability of WGAST~\cite{bouaziz2025wgast} in four regions: Rome (Italy), Cairo (Egypt), Madrid (Spain), and Montpellier (France).}
\small
\renewcommand{\arraystretch}{1.2}
\begin{tabular}{lc}
\toprule
\textbf{Region} & \textbf{Target Dates} \\
\midrule
Rome & \begin{tabular}[t]{@{}c@{}} 01 Mar 2021, 25 Jan 2022, 04 Apr 2024, \\10 Jun 2025 \end{tabular}\\
Cairo & 13 Mar 2025, 26 Jul 2025, 20 Aug 2025 \\
Madrid & 18 Jul 2025, 03 Aug 2025 \\
Montpellier & 01 Apr 2025, 22 Jul 2025 \\
\bottomrule
\end{tabular}
\label{tab:target_dates}
\end{table}


\subsubsection{Implementation Details} The weighting coefficients in Eq.~\ref{eq:ltta} are fixed to $\lambda_1 = 0.65$, $\lambda_2 = 0.30$, and $\lambda_3 = 0.25$. Landsat 8 inputs are processed using patches of size $32 \times 32$ with a stride of 8. Epistemic uncertainty is estimated using $N = 10$ MC dropout samples, for a trade-off between estimation accuracy and computational efficiency. TTA is performed for 10 epochs with a learning rate of $4 \times 10^{-4}$. All experiments are conducted on an NVIDIA RTX A6000 GPU.

\subsection{Loss Curve Analysis}

\begin{figure}[h]
  \centering
  \includegraphics[width=0.45\textwidth]
  {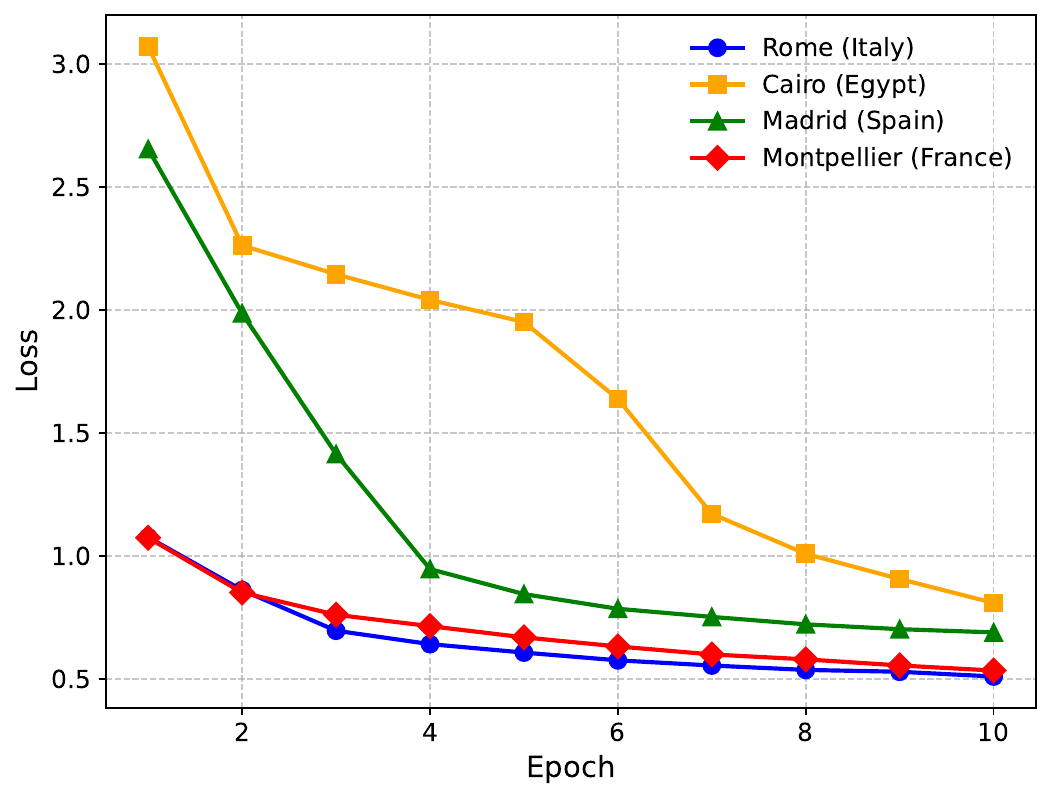}
\caption{TTA loss curves for WGAST~\cite{bouaziz2025wgast} over 10 TTA epochs in each target region. The curves show how the loss decreases and stabilizes during TTA.}
  \label{fig:loss_curves}
\end{figure}

Fig. \ref{fig:loss_curves} shows the evolution of the proposed loss over 10 TTA epochs for each target region. For all regions, the loss consistently decreases and stabilizes toward the final epochs, which indicates stable and convergent adaptation behavior. Regions that differ more strongly from the source domain, such as Cairo and Madrid, present higher initial loss values, reflecting larger domain shifts. Despite this, the loss converges within a few adaptation epochs.

\subsection{Quantitative Results}
We evaluate the proposed TTA method using Root Mean Square Error (RMSE) and Mean Absolute Error (MAE). We followed the evaluation procedure of WGAST~\cite{bouaziz2025wgast} by averaging the predicted 10 m LSTs to a 30 m resolution and comparing them with the Landsat 8 LSTs. Since this work represents the first attempt to apply TTA to a regression task in RS, we compare our approach against the pre-trained WGAST model without TTA. Table~\ref{tab:quantitative_results} presents the average results for each region over the selected target dates before and after TTA. In Rome, RMSE decreases from $3.081$ to $2.088$, corresponding to a $32.2$\% improvement, while MAE decreases from $2.735$ to $1.675$, a $38.8$\% improvement. A similar trend is observed in Cairo, with RMSE and MAE improving by 19.8\% and 19.9\%, respectively. In Madrid, RMSE and MAE improve by 27.3\% and 31.8\%, and in Montpellier by 15.8\% and 19.0\%. On average across all regions, the proposed TTA method reduces RMSE by 24.2\% and MAE by 27.9\%, demonstrating consistent performance gains under cross-region TTA scenarios.

\setlength{\tabcolsep}{6pt}
\begin{table}[h]
\centering
\caption{Quantitative evaluation of WGAST~\cite{bouaziz2025wgast} before and after applying the proposed TTA across the selected target dates in each region. RMSE and MAE values are reported for each region, with percentage improvements after TTA shown in parentheses. }
\small
\renewcommand{\arraystretch}{1.1}
\begin{tabular}{lccc}
\toprule
\textbf{Region} & \textbf{Metric} & \textbf{Before TTA} & \textbf{After TTA} \\
\midrule
\multirow{2}{*}{Rome} & RMSE ($\downarrow$) & 3.081 & \textbf{2.088 (32.23\%)} \\
                       & MAE  ($\downarrow$) & 2.735 & \textbf{1.675 (38.76\%)} \\

\midrule
\multirow{2}{*}{Cairo} & RMSE ($\downarrow$) & 3.463 & \textbf{2.778 (19.78\%)} \\
                       & MAE  ($\downarrow$) & 2.926 & \textbf{2.344 (19.89\%)} \\

\midrule
\multirow{2}{*}{Madrid} & RMSE ($\downarrow$) & 2.774 & \textbf{2.017 (27.29\%)} \\
                       & MAE  ($\downarrow$) & 2.578 & \textbf{1.758 (31.81\%)} \\

\midrule
\multirow{2}{*}{Montpellier} & RMSE ($\downarrow$) & 2.142 & \textbf{1.804 (15.78\%)} \\
                       & MAE  ($\downarrow$) & 1.800 & \textbf{1.458 (19.00\%)} \\

\midrule
\multirow{2}{*}{Average} & RMSE ($\downarrow$) & 2.865 & \textbf{2.172 (24.19\%)} \\
                       & MAE  ($\downarrow$) & 2.510 & \textbf{1.809 (27.93\%)} \\
                                             
\bottomrule
\end{tabular}
\label{tab:quantitative_results}
\end{table}

The results demonstrate that the proposed TTA method achieves performance gains even under limited target data, with only 4 target dates for Rome, 3 for Cairo, and 2 each for Madrid and Montpellier. TTA is performed over a limited number of $10$ epochs, updating only the fusion module while keeping the encoder and decoder frozen. This strategy, combined with the uncertainty-aware, LULC, and bias consistency losses, allows the model to effectively adjust to new regions without requiring label data or extensive retraining.

\section{Conclusion}
In this paper, we have proposed an uncertainty-aware TTA framework for the regression task of STF for LST estimation. Our method effectively adapts a model trained on one region to unseen target geographic regions without requiring labeled data or extensive retraining. We introduce a loss function that combines uncertainty estimation, LULC consistency correlation between LST and LULC, and bias consistency between LST at different spatial scales. The adaptation is performed by updating only the fusion module of the pre-trained STF model while keeping the encoder and decoder frozen. Experiments on four target regions with diverse climates, namely Rome in Italy, Cairo in Egypt, Madrid in Spain, and Montpellier in France, demonstrate consistent improvements in RMSE and MAE for a model pre-trained in Orléans, France, achieving average gains of 24.2\% and 27.9\%, even with limited unlabeled target data and only $10$ TTA epochs.

\vspace{0.2em}

Future work will explore extending this uncertainty-aware TTA framework to a wider range of regression-based RS tasks, including spatio-temporal prediction, environmental monitoring, and other geophysical parameter estimation problems.

\small
\bibliographystyle{IEEEtranN}
\bibliography{references}

\end{document}